\documentclass[runningheads]{llncs}

\usepackage{eccv}

\newcommand{\aitodsecondtext}[1]{\textcolor[rgb]{1,0,0}{#1}}

\newcommand{\best}[0]{\cellcolor{red!15}}
\newcommand{\second}[0]{\cellcolor[rgb]{1,1,1}}
\newcommand{\setrowcolor}[0]{\rowcolor[HTML]{EFEFEF}}

\usepackage{eccvabbrv}

\usepackage{graphicx}
\usepackage{booktabs}

\usepackage[accsupp]{axessibility}  

\usepackage{float}
\usepackage{microtype}
\usepackage{xcolor}
\usepackage{colortbl}

\usepackage{wrapfig}

\usepackage{hyperref}

\usepackage{orcidlink}

\begin{document}
\title{Visible and Clear: Finding Tiny Objects in Difference Map} 


\author{Bing Cao \and
Haiyu Yao \and
Pengfei Zhu\thanks{Corresponding author} \and
Qinghua Hu}

\authorrunning{B.~Cao et al.}

\institute{College of Intelligence and Computing, Tianjin University, Tianjin, China\newline
Tianjin Key Lab of Machine Learning, Tianjin, China\newline
\email{\{caobing,yaohaiyu,zhupengfei,huqinghua\}@tju.edu.cn}}

\maketitle

\begin{abstract}
  Tiny object detection is one of the key challenges for most generic detectors. The main difficulty lies in extracting effective features of tiny objects. Existing methods usually perform generation-based feature enhancement, which is seriously affected by spurious textures and artifacts, making it difficult to make the tiny-object-specific features visible and clear for detection. To address this issue, we propose a self-reconstructed tiny object detection (SR-TOD) framework. We for the first time introduce a self-reconstruction mechanism in the detection model, and discover the strong correlation between it and the tiny objects. Specifically, we impose a reconstruction head in-between the neck of a detector, constructing a difference map of the reconstructed image and the input, which shows high sensitivity to tiny objects. This inspires us to enhance the weak representations of tiny objects under the guidance of the difference maps. Thus, improving the \textit{visibility} of tiny objects for the detectors. Building on this, we further develop a Difference Map Guided Feature Enhancement (DGFE) module to make the tiny feature representation more \textit{clear}. In addition, we further propose a new multi-instance anti-UAV dataset. Extensive experiments demonstrate our effectiveness. The code is available: \url{https://github.com/Hiyuur/SR-TOD}.
\end{abstract}

\section{Introduction}\label{Introduction}
\label{sec:intro}
The subfield of object detection that identifies and categorizes objects with diminutive dimensions is known as tiny object detection.
According to setting in MS COCO~\cite{lin2014microsoft}, an object qualifies as \textit{small} if it occupies an area equal to or less than 32×32 pixels.
The AI-TOD benchmark~\cite{wang2021tiny} refines this definition and categorizes objects as ``very tiny'' if they span 2 to 8 pixels, ``tiny'' for 8 to 16 pixels, and ``small'' for 16 to 32 pixels. We term these objects uniformly as ``\textit{tiny}'' objects.
Tiny objects commonly appear in a variety of real-world applications, such as autonomous driving, industrial inspection, and pedestrian detection, often restricted by considerable imaging distances or the inherently minute size of the objects.  
Despite its relevance, tiny object detection (TOD) remains an arduous endeavor, with even state-of-the-art detectors struggling to bridge the performance disparity between tiny and normal-sized object detection~\cite{cheng2023towards}.
The pursuit of improved methods in this area is of considerable theoretical and practical importance.

\begin{figure}[t]
     \centering
     \includegraphics[width=1\linewidth]{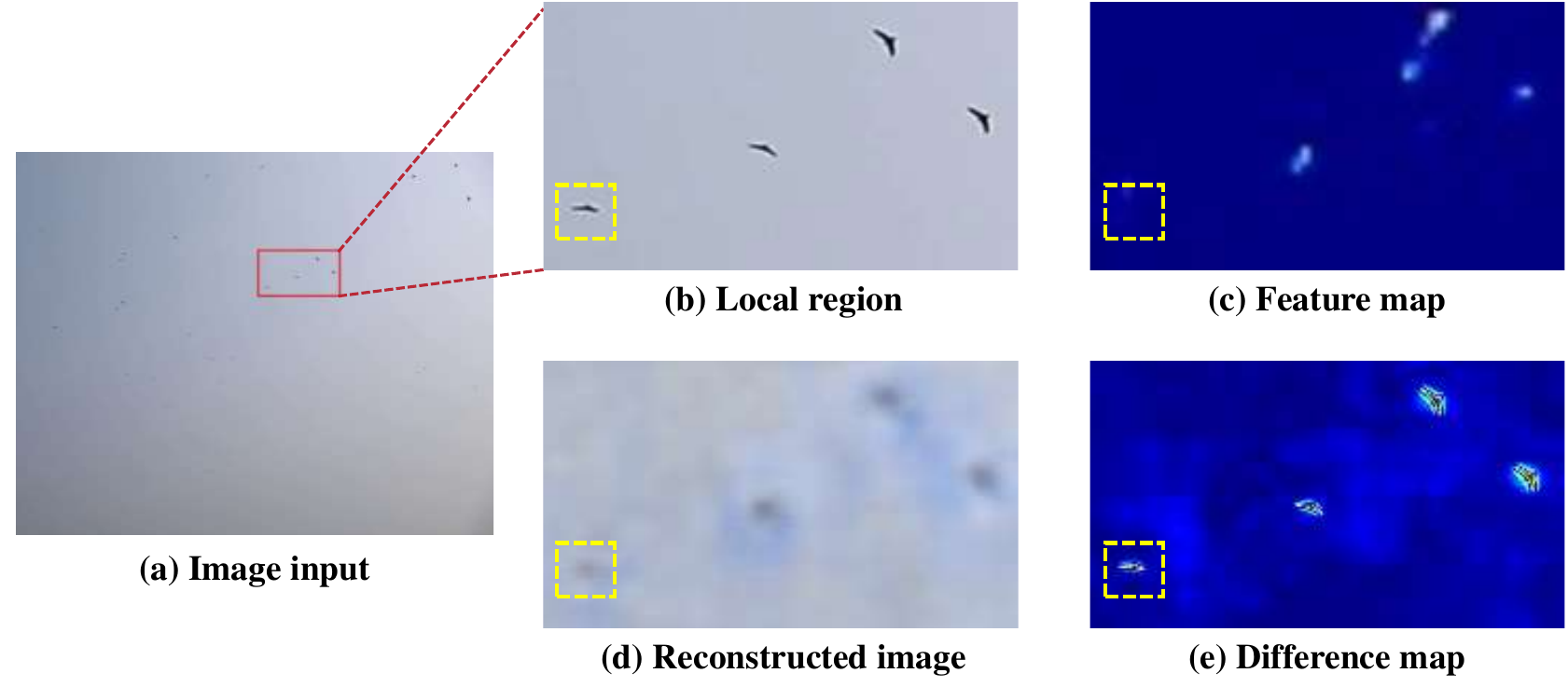}
     \caption{Visualization of the image self-reconstruction mechanism. The results are from Cascade R-CNN~\cite{cai2018cascade}. \textbf{a} shows the entire image. To demonstrate the effect more clearly, \textbf{b} zooms in on the local region in the red box. \textbf{c} shows the visual heatmap of the corresponding region in the feature map of FPN P2. \textbf{d} is the reconstructed image. \textbf{e} is the difference map. The yellow dotted box highlights the tiny drone whose signal is almost wiped out in the feature map.}
     \label{fig:fig1}
\end{figure}

Relative to prevalent complications in general object detection, such as object occlusion, tiny object detection presents some unique challenges.
The foremost challenge arises from the issue of information loss on tiny objects~\cite{cheng2023towards}.
Prevailing detection architectures employ backbone networks, like ResNet and others~\cite{he2016deep,xie2017aggregated,gao2019res2net}, for feature extraction. Nonetheless, these frameworks often implement downsampling operations aimed at eliminating noisy activations and diminishing the spatial resolution of feature maps, a process that unavoidably results in the information loss of tiny objects.
In addition, TOD is also impeded by the inherently limited size and scant information content of tiny objects, resulting in substantial information loss throughout the feature extraction phase.
Such degradation in object representation significantly hampers the detection head's capacity to localize and distinguish between tiny objects. Consequently, tiny objects become indiscernible to the detector. 
In particular, the weak signals of ``very tiny'' objects are almost wiped out under these conditions, making it difficult for detectors to localize and identify them.
As shown in Fig.~\ref{fig:fig1}, the feature heatmaps display the activated signals of the detection model towards tiny drones, which is often weak for tiny objects. For instance, the tiny object in the lower left almost disappears, affecting the detection performance.
This indicates that many tiny objects are insufficiently visible to detectors.
Therefore, the performance of generic detectors dramatically decreases in tiny object detection tasks~\cite{wang2021tiny,yu2020scale}.

Many existing methods~\cite{rabbi2020small,bashir2021small,bai2018sod,bai2018finding,noh2019better,li2017perceptual} often use a generation approach with super-resolution architecture to alleviate the low-quality representation issue of tiny objects caused by information loss. These methods typically incorporate generative adversarial networks~\cite{goodfellow2014generative} into the object detection framework, constructing pairs of high-resolution and low-resolution samples. This approach enables the generator to learn to restore the distorted structure of tiny objects, aiming to enhance the features of low-quality tiny objects. Nonetheless, these approaches typically necessitate a substantial volume of medium and large-sized samples, posing a significant challenge in executing super-resolution on tiny objects with weak signals. Furthermore, these methods are inclined to fabricate spurious textures and artifacts, which decreases detection performance~\cite{deng2021extended}. It is worth noting that super-resolution architecture brings a significant amount of computational overhead, complicating the end-to-end optimization~\cite{cheng2023towards}.

Compared to the inefficient feature enhancement with intricate super-resolution architectures, recapturing lost information in the backbone network is a more intuitive and reasonable strategy.
For the first time, we introduce a simple but effective image self-reconstruction mechanism into the object detection framework. The feature maps extracted by the detection model are restored by a reconstruction head, which is constrained by mean square error at the pixel level.
Note that, image reconstruction, a task situated within the low-level vision domain, is highly sensitive to pixel changes~\cite{cao2023autoencoder}. 
Since we reconstruct the input image from the detection model, the image regions that are difficult to restore may correspond to those where structural and textural information is severely lost during the feature extraction of the backbone network, notably tiny objects.
With the disparities between the reconstructed and the original images, we can pinpoint the regions that have undergone significant information loss, which in turn provides potential prior knowledge for detecting tiny objects.
Therefore, we subtract the original image from the self-reconstructed image to construct a difference map, as shown in Fig.~\ref{fig:fig1}.
We first discover a strong correlation between the self-reconstructed difference map and tiny objects.
The ``\textit{very tiny}'' objects that are almost eliminated in Fig.~\ref{fig:fig1} can also be clearly shown in the difference map.
Most tiny objects in the image have a significant activation in the difference map. Moreover, the difference map also preserves the main structures of the tiny objects.
We believe that the difference map manifests the detector's pixel-level discernment for the areas of interest, as well as the potential location and structure of the tiny objects.
Overall, the difference map enables tiny objects that have weak signals to be more \textit{visible}.

Therefore, building upon this discovery, we further incorporate the prior information from the difference map into the object detection model.
We develop a simple and effective Difference Map Guided Feature Enhancement module, which calculates an element-wise attention matrix by reweighting the difference map along the channel dimension to perform feature enhancement for tiny objects. Consequently, by converting the reconstruction loss into a constraint specifically targeting tiny objects, we enhance the model's capacity for detecting such objects, making the tiny object more \textit{clear} to detectors.

Furthermore, we collected a new anti-UAV dataset, named DroneSwarms, which is a typical tiny object detection scenario under various complex backgrounds and lighting conditions. Our DroneSwarms has the smallest average size (about 7.9 pixels) of drones for anti-UAV. Experiments on our dataset and two other datasets demonstrate our superiority against the competing methods.

The contributions of this paper can be summarized as follows:
\begin{itemize}
\item[(1)]We propose a self-reconstructed tiny object detection (SR-TOD) framework, revealing for the first time the robust association between the difference map and tiny objects, thus offering prior information on the position and structure of tiny objects. We effectively convert the typically lost information of tiny objects into actionable prior guidance.
\item[(2)]We design a Difference Map Guided Feature Enhancement (DGFE) module, which improves the feature representation of tiny objects to make them clearer.
The DGFE module can be easily and flexibly integrated into generic detectors, effectively improving the performance of tiny object detection.
\item[(3)]We propose a new tiny object detection dataset for anti-UAV, named DroneSwarms, which has the smallest average object size currently. Extensive experiments on our dataset and two other datasets with a large number of tiny objects validate our effectiveness against the competing methods.
\end{itemize}

\section{Related Work}

\textbf{Object detection} algorithms come in various types.
Two-stage detectors feed the extracted feature maps into a Region Proposal Network to extract proposals in the first stage. Then, in the second stage, they perform classification and regression tasks based on these proposals, providing high recognition and localization accuracy.
Classic two-stage detectors include Fast R-CNN~\cite{girshick2015fast}, Faster R-CNN~\cite{ren2015faster}, Cascade R-CNN~\cite{cai2018cascade} \etc.
One-stage detectors directly perform object localization and classification on input images, offering faster processing speeds.
Representative examples of one-stage detectors include the YOLO series~\cite{bochkovskiy2020yolov4,redmon2017yolo9000,redmon2018yolov3} and RetinaNet~\cite{lin2017focal}.
Besides, anchor-free algorithms such as FCOS~\cite{tian2019fcos} and FoveaBox~\cite{kong2020foveabox} predict objects based on central points, while methods like CornerNet~\cite{law2018cornernet}, Grid R-CNN~\cite{lu2019grid}, and RepPoints~\cite{yang2019reppoints} make predictions based on keypoints.
Recent detectors such as DETR~\cite{carion2020end}, Deformable DETR~\cite{zhu2020deformable}, and Sparse R-CNN~\cite{sun2021sparse} explore new paradigms for end-to-end object detection.
These algorithms commonly use FPN~\cite{lin2017feature} as the neck module, which allows our method to be easily integrated into most generic detectors.

\noindent\textbf{Tiny Object Detection} is a challenging task for most generic detectors.
In recent years, the research in tiny object detection has mainly focused on data augmentation, scale awareness, context modeling, feature imitation, and label assignment~\cite{cheng2023towards}. \textit{Data augmentation}:
Krisantal \etal.~\cite{kisantal2019augmentation} copy and paste tiny objects to increase the number of samples. DS-GAN~\cite{bosquet2023full} designs a new data augmentation pipeline to generate high-quality synthetic data for tiny objects.
\textit{Scale awareness}: Lin~\cite{lin2017feature} \etal. propose the most popular multi-scale network feature pyramid network using a pyramid of feature levels and feature fusion. Singh \etal.~\cite{singh2018analysis} design the Scale Normalization for Image Pyramids (SNIP) to select some instances for training.
PANet~\cite{liu2018path} enriches the feature hierarchy through bidirectional paths and enhanced deeper features with accurate positioning signals.
NAS-FPN~\cite{ghiasi2019fpn}, Bi-FPN~\cite{tan2020efficientdet} and Recursive-FPN~\cite{qiao2021detectors} further develop upon the basis of FPN.
Gong \etal. ~\cite{gong2021effective} adjust the coupling between adjacent layers of FPN by setting fusion factors to optimize feature fusion. Yang \etal. ~\cite{yang2022querydet} design a cascaded sparse query mechanism to effectively utilize high-resolution features to enhance the detection performance of tiny objects while maintaining fast inference speed.
\textit{Context modeling}:
Chen \etal.~\cite{chen2017r} utilizes context region representations containing proposed patches for subsequent recognition.
SINet~\cite{hu2018sinet} introduces a context-aware RoI pooling layer to maintain contextual information.
\textit{Feature imitation}:
Many methods~\cite{bai2018sod,bai2018finding,li2017perceptual}utilize generative adversarial networks to perform super-resolution on tiny objects.
Noh \etal.~\cite{noh2019better} alleviated the mismatch between the receptive fields of high-resolution features and low-resolution features through dilated convolutions.
Deng \etal.~\cite{deng2021extended} proposed a feature texture transfer module to expand the feature pyramid, enabling the new feature layers to contain more detailed information of tiny objects.
\textit{Label assignment}:
ATSS~\cite{zhang2020bridging} adaptively adjusts positive and negative samples based on their statistical features.
Xu \etal. ~\cite{xu2022rfla} propose a simple yet effective strategy called Receptive Field-based Label Assignment (RFLA) to alleviate the scale-sample imbalance problem in anchor-based and anchor-free detectors.

Many tiny object detection methods have not focused on the key issue of information loss.
Feature imitation methods attempt to alleviate this issue through generation but tend to fabricate spurious textures and artifacts~\cite{cheng2023towards,deng2021extended}.
In contrast, we introduce an image self-reconstruction mechanism to identify areas of significant information loss and leverage this prior knowledge to improve tiny object detection performance.

\noindent\textbf{Anti-UAV Dataset.}
Drones, due to their tiny size and low cost among other features, are widely used in inspection and surveillance.
However, illegal drone flights have introduced many potential risks and threaten public safety, making anti-UAV measures an important task. 
Since drones are tiny in size and often fly at medium to high altitudes, they tend to appear very small in images.
Therefore, anti-UAV technology is very well suited for the application of tiny object detection.

To our knowledge, there are three publicly available visible light anti-UAV datasets in the current field.
MAV-VID~\cite{rodriguez2020adaptive} consists of 64 video sequences containing a single drone captured from multiple viewpoints, with the targets predominantly located in the central region of the images.
The drones are relatively large in scale, with an average size of approximately 166 pixels. Drone-vs-Bird~\cite{coluccia2021drone} consists of 77 video sequences and is used to distinguish between drone and bird targets.
The average size of the drones in the dataset is approximately 28 pixels.
DUT Anti-UAV~\cite{zhao2022vision} is divided into two subsets: detection and tracking.
The majority of the images in this dataset contain single objects and almost no very tiny objects. 


\section{Method}

\subsection{Overall Architecture}

\begin{figure}[t]
    \centering
    \includegraphics[width=12cm]{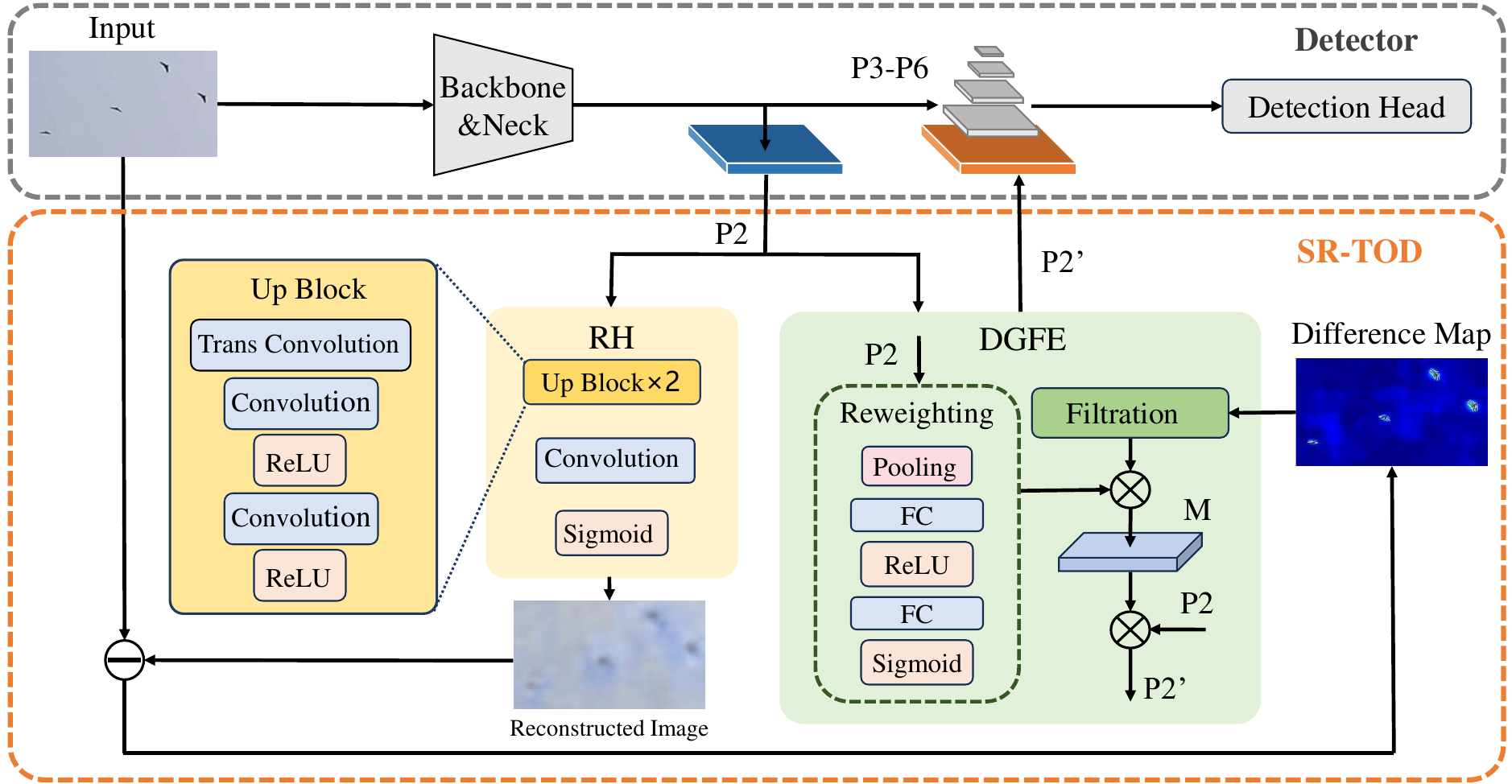}
    \caption{Overall model architecture. RH refers to the reconstruction head, DGFE is the difference map guided feature enhancement. $P2-P6$ denote the feature maps extracted from FPN. $P2'$ is the enhanced feature map. $M$ is the element-wise attention matrix.}

    \label{fig:framework}
\end{figure}

In this paper, we propose a tiny object detection framework based on the image self-reconstruction mechanism, as shown in Fig. \ref{fig:framework}, to address the major challenge of severe information loss in the feature extraction process of the backbone network.
Initially, the image is fed into the backbone network, which extracts features from this input and relays the feature maps to the neck module—typically FPN~\cite{lin2017feature}—to create a multi-scale feature pyramid ranging from $P2$ to $P5$.
In line with the architecture of prevalent detectors, the task of tiny object detection is designated to $P2$.
Consequently, the interfacing of our self-reconstruction mechanism with the detector occurs exclusively via $P2$.
We feed $P2$ into the reconstruction head, which will be detailed in Sec. \ref{sec3.2}.
The reconstruction head generates images that are dimensionally consistent with the original input.
By subtracting the original images from the reconstructed imges, taking the absolute values, and averaging across the three color channels, we obtain the difference map.
We feed  both the difference map and $P2$ into the Difference Map Guided Feature Enhancement (DGFE) module, which will be detailed in Sec. \ref{sec3.3}.
The DGFE module enhance specific tiny object features in $P2$ based on the prior knowledge of the difference map, resulting in $P2'$.
This enhanced feature map, $P2'$, supplants the original $P2$ as the bottom level of the feature pyramid, subsequently fed into the detection head.
Note that some one-stage detectors ~\cite{lin2017focal} rely exclusively on lower-resolution feature maps, $P3$, hence our framework can also use $P3$ for image reconstruction.
Owing to the widespread adoption of Feature Pyramid Networks (FPN)~\cite{lin2017feature} and its variants as the neck module, our framework is readily integrable with the majority of contemporary detection models.

\subsection{Difference Map}\label{sec3.2}

The downsampling process inherent to the feature extraction in the backbone network inevitably results in information loss of objects. This loss is particularly acute for tiny objects due to their restricted sizes.
In this case, the weak signals of tiny objects are almost wiped out, making it difficult for the detection head to predict from such low-quality representations~\cite{cheng2023towards}.
In response to this problem, we reconsidered the attributes of feature maps at different levels within the FPN framework.
Given that high-level, low-resolution features are imbued with rich semantic content, while low-level, high-resolution features possess greater local detail and positional information, we favor image reconstruction using the low-level feature maps.

The U-Net~\cite{ronneberger2015u} architecture, generally used for image reconstruction tasks, bears a similarity to the popular FPN module of object detection models.
Therefore, we designed a simple reconstruction head inserted into the top-down path of the FPN, as shown in Fig. \ref{fig:framework}.
Given a low-level feature map $X$ with $C$ channels and size of $H\times W$, the upsampling result $Up(X)\in \mathbb{R}^{\frac{C}{2}\times 2H \times 2W}$can be computed as
\begin{equation}
    \label{eq1}
    Up(X)=\delta(Conv2(\delta(Conv1(TranConv(X))))),
\end{equation}
where $\delta$ denotes Rectified Linear Unit (ReLU)~\cite{nair2010rectified}.
$Conv1$ and $Conv2$ denote the convolutions with kernel size of $C\times C\times 3 \times 3$.
$TranConv$ denotes Transpose Convolution~\cite{zeiler2011adaptive}.
The kernel size of $TranConv$ is $\frac{C}{2}\times C\times 4\times 4$, and the step length is 2.
Given the original image $I_o\in \mathbb{R}^{3 \times H\times W}$ and the bottom-level feature map $P2\in \mathbb{R}^{C\times \frac{H}{4} \times \frac{W}{4}}$, the reconstruction head shown in Fig. \ref{fig:framework} can be formulated as
\begin{equation}
    \label{eq2}
    I_r=\sigma(Conv(Up(Up(P2)))),
\end{equation}
where $I_r$ is the reconstructed image.
$\sigma$ denotes Sigmoid function and the kernel size of $Conv$ is $3 \times C\times 3\times 3$.
The reconstructed image $I_r$ obtained after two upsampling operations is of the same shape as the original image $I_o$.
Therefore, the difference map $D$ can be computed as
\begin{equation}
    \label{eq3}
    D=Mean_{channel}(Abs(I_r-I_o)), 
\end{equation}
where $Mean_{channel}$ denotes computing the mean value along the channel dimension, and $Abs$ denotes computing the absolute value of each element.

The optimization of the reconstruction head's parameters is achieved by computing the Mean Squared Error (MSE) loss between the original and reconstructed images.

\subsection{Difference Map Guided Feature Enhancement}\label{sec3.3}

Following the creation of the difference map via the self-reconstruction mechanism, a critical challenge is the effective utilization of the difference map's prior information to bolster tiny object detection capabilities.
Considering that the difference map represents the potential location and structural information of tiny objects, we designed a simple plug-and-play module called Difference Map Guided Feature Enhancement (DGFE).
The DGFE module calculates an element-wise attention matrix, denoted as $M\in \mathbb{R}^{C\times H \times W}$, from the difference map with the aim of executing targeted feature enhancement of tiny objects within $P2\in \mathbb{R}^{C\times H \times W}$, as shown in Fig. \ref{fig:framework}.
\\ \textbf{Filtration.}
The inherent discrepancies between the reconstructed and original images result in activation across virtually the entire difference map to varying extents. 
To filter out most of the noise signals and make the difference map more sharp, we construct a binary difference map $D_b\in \mathbb{R}^{1\times 4H \times 4W}$.
For this purpose, we set a learnable threshold $t$. 
Given a difference map $D\in \mathbb{R}^{1\times 4H \times 4W}$ with the same size as the original image and considering the back propagation of gradients, the $Filtration(D)\in \mathbb{R}^{1\times H \times W}$ in Fig. \ref{fig:framework} can be formulated as
\begin{equation}
    \label{eq4}
    Filtration(D)=Resize(D_b)+1=Resize((Sign(D-t)+1)\times 0.5)+1, 
\end{equation}
where $Sign$ denotes the Sign function, and Resize denotes resizing $D_b$ to the same size as $P2$.
Specifically, $Resize(D_b)+1$ can preserve the valuable information originally present in feature map $P2$, ensuring that it is not affected by regions in $D_b$ where the values are 0.
\\ \textbf{Reweighting.}
Since the difference map only contains spatial information, we need to utilize broadcasting to reweight it along the channel dimension, which helps to sustain the feature diversity.
Given the feature map $P2\in \mathbb{R}^{C\times H \times W}$, weights $Reweighting(P2)\in \mathbb{R}^{C\times 1 \times 1}$ can be computed as
\begin{equation}
    \label{eq5}
    Reweighting(P2)=\sigma(MLP(AvgPool(P2))+MLP(MaxPool(P2))), 
\end{equation}
where, $AvgPool$ denotes the average pooling along the spatial dimension, and $MaxPool$ denotes the max pooling.
The $MLP$ includes two fully connected layers and a ReLU function.
Therefore, the DGFE module can be formulated as
\begin{equation}
    \label{eq6}
    P2'= M \otimes P2=(Reweighting(P2) \otimes Filtration(D)) \otimes P2, 
\end{equation}
where $M\in \mathbb{R}^{C\times H \times W}$ denotes the element-wise attention matrix, and $P2'\in \mathbb{R}^{C\times H \times W}$ denotes the feature map that has been specifically enhanced for tiny objects.



\section{Experiment}

\subsection{Experimental Setting}
\textbf{Datasets.}
The experiments are conducted on three datasets.
The VisDrone2019~\cite{du2019visdrone} encompasses objects spanning 10 categories, and all images are captured from the perspective of drones.
The AI-TOD dataset~\cite{wang2021tiny} comprises objects across 8 categories, and the images are collected from various datasets containing tiny objects. Our anti-UAV dataset DroneSwarms has the smallest average absolute object size of about 7.9 pixels. 
All selected datasets contain a large number of tiny objects, especially tiny objects with sizes below 16 pixels.

\begin{table}[tb]
  \caption{Results on DroneSwarms. RFLA is based on Cascade R-CNN. The best results are highlighted. $\Delta$ represents the improvement of the best towards its baseline.
  }
  \label{tab:main_results_DroneSwarms}
  \centering
  \setlength{\tabcolsep}{6pt}
  \begin{tabular}{l|ccc|ccc}
    \toprule
\setrowcolor    Method  & AP & AP$_{0.5}$ & AP$_{0.75}$ & AP$_{vt}$ & AP$_t$ & AP$_s$ \\
    \midrule
    TridentNet~\cite{li2019scale} & 7.1 & 14.8 & 6.0 & 0.1 & 12.2 & 56.6 \\
    ATSS~\cite{zhang2020bridging} & 26.2 & 67.0 & 14.4 & 16.5 & 39.1 & 57.7 \\
    CZ Det~\cite{meethal2023cascaded} & 26.5 & 72.6 & 11.7 & - & - & - \\
    DyHead~\cite{dai2021dynamic} & 26.6 & 68.2 & 14.7 & 16.8 & 39.8 & 57.7 \\
    RetinaNet~\cite{lin2017focal} & 27.3 & 75.8 & 11.3 & 21.0 & 40.5 & 56.2 \\
    CFINet~\cite{yuan2023small} & 33.7 & 79.3 & 21.1 & 26.0 & 44.3 & 55.6 \\
    Faster R-CNN~\cite{ren2015faster} & 35.0 & 83.9 & 20.9 & 27.3 & 44.3 & 56.5 \\
    Cascade RPN~\cite{vu2019cascade} & 35.3 & 83.1 & 22.5 & 27.1 & 45.3 & 58.2 \\
    Cascade R-CNN~\cite{cai2018cascade} & 36.4 & 85.0 & 23.5 & 28.8 & 45.7 & 58.3 \\
    RFLA~\cite{xu2022rfla} & 36.9 & 86.3 & 23.4 & 29.5 & 45.3 & 58.0 \\
    DetectoRS~\cite{qiao2021detectors} & 37.9 & 87.4 & 24.8 & 30.5 & 46.9 & \second{}59.3 \\
    \midrule
    Deformable-DETR~\cite{zhu2020deformable} & 32.6 & 81.9 & 17.3 & 25.3 & 42.3 & 55.8 \\
    DINO~\cite{zhang2022dino} & 35.4 & 85.9 & 20.3 & 28.8 & 44.7 & 57.3 \\
    \midrule
    DINO w/ SR-TOD & 35.6 & 86.0 & 20.7 & 28.4 & 44.8 & 58.2 \\
    \midrule
    RetinaNet w/ SR-TOD & 28.7 & 77.2 & 12.9 & 22.3 & 42.1 & 56.9 \\
    Faster R-CNN w/ SR-TOD & 36.3 & 86.4 & 21.7 & 29.5 & 45.1 & 56.0 \\
    Cascade R-CNN w/ SR-TOD & 38.3 & 87.4 & 25.4 & 30.8 & 47.4 & \best\textbf{59.4} \\
    DetectoRS w/ SR-TOD & \second{}38.8 & \second{}87.9 & \best\textbf{26.3} & \second{}31.6 & \best\textbf{47.7} & 59.0 \\
    RFLA w/ SR-TOD & \best\textbf{39.0} & \best\textbf{88.9} & \second{}25.8 & \best\textbf{31.8} & \second{}47.6 & 59.2 \\
    \midrule
    $\Delta$ Improvement & \textcolor{blue}{2.1} & \textcolor{blue}{2.6} & \textcolor{blue}{1.5} & \textcolor{blue}{2.3} & \textcolor{blue}{0.8} & \textcolor{blue}{1.1} \\
  \bottomrule
  \end{tabular}
\end{table}

\noindent\textbf{Implementation Details.}
All core codes are built upon MMdetection~\cite{chen2019mmdetection}.
Since the DroneSwarms dataset mainly consists of a large number of tiny objects, in order to ensure model convergence, we set the initial learning rate to 0.0025 for all experiments on the DroneSwarms dataset, and use the Stochastic Gradient Descent (SGD) optimizer training models for 20 epochs with 0.9 momenta, 0.0001 weight decay, 2 batch size, and 2 anchor scale.
For experiments on VisDrone2019~\cite{du2019visdrone} and AI-TOD~\cite{wang2021tiny}, we strictly followed all the experiment settings of RFLA~\cite{xu2022rfla} on these two datasets, such as using the SGD optimizer for 12 epochs training.
And the initial learning rate is set to 0.005 and decays at the 8th and 11th epochs.
The ImageNet~\cite{russakovsky2015imagenet} pre-trained model, ResNet-50, is used as the backbone.
All the other parameters of baselines are set the same as default in MMdetection.
Our experiments are conducted on Huawei Atlas 800 Training Server with CANN and NVIDIA RTX 3090 GPU.

\noindent \textbf{Evaluation Metrics.}
While the Average Precision (AP) evaluation metric from MS COCO~\cite{lin2014microsoft} is prevalent among object detection algorithms, it is developed only with generic detectors in mind.
This metric broadly categorizes objects with size under 32 pixels as small, employing only the AP$_s$ for their assessment.
Thus, to more effectively demonstrate our method's performance in detecting tiny objects of diverse sizes, the evaluation metric follows AI-TOD benchmark~\cite{wang2021tiny}.
Note that AP$_{vt}$,AP$_t$, AP$_s$ are APs for very tiny, tiny, small scales, respectively.
The definitions of these scales are introduced in Sec. \ref{Introduction}.

\begin{table}[tb]
  \caption{Results on VisDrone2019. The best results are highlighted.
  }
  \label{tab:main_results_VisDrone2019}
  \setlength{\tabcolsep}{6pt}
  \centering
  \begin{tabular}{l|ccc|ccc}
    \toprule
    \setrowcolor Method  & AP & AP$_{0.5}$ & AP$_{0.75}$ & AP$_{vt}$ & AP$_t$ & AP$_s$ \\
    \midrule
    Faster R-CNN~\cite{ren2015faster} & 23.9 & 42.2 & 23.8 & 0.1 & 6.5 & 21.1 \\
    Cascade R-CNN~\cite{cai2018cascade} & 25.2 & 42.6 & 25.9 & 0.1 & 7.0 & 22.5 \\
    DetectoRS~\cite{qiao2021detectors} & 26.3 & 43.9 & 26.9 & 0.1 & 7.5 & 23.3 \\
    RFLA~\cite{xu2022rfla} & 27.2 & \second{}48.0 & 26.6 & \second{}4.5 & \second{}13.0 & 23.6 \\
    \midrule
    Faster R-CNN w/ SR-TOD & 26.3 & 46.8 & 26.0 & 2.9 & 11.0 & 23.7 \\
    DetectoRS w/ SR-TOD & 27.2 & 47.1 & \second{}27.2 & 2.4 & 11.7 & 24.2 \\
    Cascade R-CNN w/ SR-TOD & \second{}27.3 & 46.9 & \best\textbf{27.5} & 2.3 & 11.5 & \best\textbf{24.7} \\
    RFLA w/ SR-TOD & \best\textbf{27.8} & \best\textbf{48.8} & \textbf{27.5} & \best\textbf{4.8} & \best\textbf{13.2} & \second{}24.5\\
    \midrule
    $\Delta$ Improvement & \textcolor{blue}{0.6} & \textcolor{blue}{0.8} & \textcolor{blue}{$1.6$} & \textcolor{blue}{0.3} & \textcolor{blue}{0.2} & \textcolor{blue}{2.2} \\
  \bottomrule
  \end{tabular}
\end{table}

\subsection{Results on DroneSwarms}

 As shown in Tab.~\ref{tab:main_results_DroneSwarms}, RFLA w/ SR-TOD achieves an AP of 39.0, surpassing the competing methods by 1.1 AP and resulting in an overall performance boost of 2.1 AP.
 Notably, RFLA w/ SR-TOD registers a substantial performance gain of 2.3 points in both AP$_{vt}$ and AP$_t$, underscoring the pronounced efficacy of our method in detecting tiny objects.
 Furthermore, within the Cascade R-CNN framework, Cascade R-CNN with SR-TOD exceeds RFLA by 1.4 AP, indicating a notable improvement of 1.9 AP.
 Moreover, our method has also manifested clear performance gains when applied to other detectors.
 One-stage detectors typically perform worse than multi-stage detectors due to the lack of multi-stage regression~\cite{yu2020scale,zhu2018visdrone,du2019visdrone,xu2022rfla}.
 Even though the one-stage detector RetinaNet~\cite{lin2017focal} only utilizes the feature map $P3$ with lower resolution from FPN, our method achieves a 1.4 AP improvement.
 It is particularly noteworthy that, due to focusing on different challenges of tiny object detection, our framework is qualified to work with other methods like RFLA.
 Our framework does not conflict with methods like DetectoRS~\cite{qiao2021detectors}, which also enhance feature representation by improving FPN. Compared to the baseline methods, our method achieves consistent improvements.
We have also conducted experiments with transformer-based detectors on the DroneSwarms dataset. 
Due to the slow convergence speed of transformer-based detectors, the schedule for Deformable-DETR~\cite{zhu2020deformable} is set to 50 epochs as recommended.
Obviously, our method shows significant superiority, even compared with the advanced transformer-based detector DINO~\cite{zhang2022dino}. 
Besides, we have also integrated our SR-TOD with DINO, which introduces improvements of 0.2 AP and 0.9 AP$_s$.
This further indicates our effectiveness in finding tiny objects when integrating with different architectures, including transformer and CNN. 
In addition, these experiments that transformer-based detectors shows limited performance in detecting tiny objects. 
Recent studies also indicated that DETR lacks effective multi-scale features and that the sparse query paradigm does not adequately cover tiny objects~\cite{carion2020end,cheng2023towards}. Moreover, the cost of using high-resolution feature maps beneficial for tiny objects in Transformer-based detectors is prohibitively high due to attention mechanisms~\cite{zhu2020deformable}.

\subsection{Results on VisDrone2019 and AI-TOD}



\begin{table}[tb]

  \caption{Results on AI-TOD. The best and second results are highlighted.
  }
  \label{tab:main_results_AI-TOD}
  \setlength{\tabcolsep}{6pt}
  \centering
  \begin{tabular}{l|ccc|ccc}
    \toprule
 \setrowcolor   Method  & AP & AP$_{0.5}$ & AP$_{0.75}$ & AP$_{vt}$ & AP$_t$ & AP$_s$ \\
    \midrule
    Faster R-CNN~\cite{ren2015faster} & 11.7 & 27.4 & 8.2 & 0.0 & 8.6 & 23.7 \\
    Cascade R-CNN~\cite{cai2018cascade} & 14.0 & 31.2 & 10.7 & 0.1 & 10.3 & 26.2 \\
    DetectoRS~\cite{qiao2021detectors} & 14.6 & 31.8 & 11.5 & 0.0 & 11.0 & \second{}27.4 \\
    RFLA~\cite{xu2022rfla} & 21.7 & 50.5 & 15.3 & 8.3 & 21.8 & 26.3 \\
    HANet~\cite{guo2023save} &  \aitodsecondtext{22.1} &  \aitodsecondtext{53.7} & 14.4 & \best\textbf{10.9} & 22.2 & 27.3 \\
    \midrule
    Faster R-CNN w/ SR-TOD & 20.6 & 49.8 & 13.2 & 7.1 & 21.3 & 26.3 \\
    RFLA w/ SR-TOD & 21.8 & 50.8 & 15.4 & 9.7 & 21.8 &  \aitodsecondtext{27.4} \\
    Cascade R-CNN w/ SR-TOD & 21.9 & 50.6 &  \aitodsecondtext{15.6} & 9.6 &  \aitodsecondtext{22.4} & 26.7 \\
    DetectoRS w/ SR-TOD & \best\textbf{24.0} & \best\textbf{54.6} & \best\textbf{17.1} &  \aitodsecondtext{10.1} & \best\textbf{24.8} & \best\textbf{29.3} \\
    \midrule
    $\Delta$ Improvement & \textcolor{blue}{9.4} & \textcolor{blue}{22.8} & \textcolor{blue}{$5.6$} & \textcolor{blue}{10.1} & \textcolor{blue}{13.8} & \textcolor{blue}{1.9} \\
  \bottomrule
  \end{tabular}
\end{table}


We further compare our model with the competing methods on VisDrone2019~\cite{du2019visdrone} and AI-TOD~\cite{wang2021tiny}, as shown in Tab. \ref{tab:main_results_VisDrone2019} and Tab. \ref{tab:main_results_AI-TOD}.

\noindent\textbf{VisDrone2019.}
RFLA w/ SR-TOD has a improvement of 0.6 AP and achieves 27.8 AP on VisDrone2019~\cite{du2019visdrone}, showing a clear advantage.
Furthermore, Cascade R-CNN w/ SR-TOD has a performance improvement of 2.1 AP, with the highest performance in AP$_s$. 
This suggests that Cascade R-CNN w/ SR-TOD demonstrates better performance on small objects exceeding 16 pixels in size.
Importantly, Cascade R-CNN w/ SR-TOD and RFLA w/ SR-TOD show significant improvements of 1.6 points and 0.9 points in AP$_{0.75}$, respectively.
The enhanced cascade structure with SR-TOD effectively leverages multi-stage regression to increase the localization accuracy for tiny objects.

\noindent \textbf{AI-TOD.}
DetectoRS w/ SR-TOD attains an AP of 24 on the AI-TOD dataset~\cite{wang2021tiny}, exceeding RFLA by a margin of 2.3 AP and significantly surpassing all competitors.
All detectors have significant performance improvements.
Significantly, Cascade R-CNN w/ SR-TOD surpasses RFLA by 1.3 points in AP$_{vt}$ and by 0.6 points in AP$_t$, underscoring the substantial impact of our method on tiny object detection within the AI-TOD dataset~\cite{wang2021tiny}.
Our method outperforms HANet~\cite{guo2023save} on most metrics (\eg, AP: $22.1 \rightarrow 24.0$, AP$_t$: $22.2 \rightarrow 24.8$), indicating our superiority.
HANet~\cite{guo2023save} aims to obtain scale-specific feature subspaces with a predicted activation map.
Although the shallow features further compensate for the detailed information of the specific scale feature space, it is still hard to capture tiny objects as their weak signals are not enhanced in shallow layers. Compared to the activation map of HANet, our self-reconstructed difference map is more sensitive to information loss, especially for tiny objects.
The consistent improvement on various datasets indicates the generality of SR-TOD.
Experiments on more datasets are provided in the supplementary materials.

\begin{table}
\begin{minipage}[c]{0.49\textwidth}

\captionof{table}{Influence of different designs. RH means Reconstruction Head. The results are on DroneSwarms. }
 
\centering
   \scalebox{1}{
\begin{tabular}{cc|cc|ccc}
    \toprule
  \setrowcolor  RH & DGFE & AP & AP$_{0.5}$ & AP$_{vt}$ & AP$_t$ & AP$_s$ \\
    \midrule
     & & 36.4 & 85.0 & 28.8 & 45.7 & 58.3 \\
    \checkmark & & 36.5 & 84.9 & 28.7 & 45.9 & 58.5 \\
    \checkmark & \checkmark & \textbf{38.3} & \textbf{87.4} & \textbf{30.8} & \textbf{47.4} & \textbf{59.4} \\
  \bottomrule
  \end{tabular}}
 \label{tab:ablation_1}
\end{minipage}
\begin{minipage}[c]{0.49\textwidth}
\centering

   \captionof{table}{Effectiveness of threshold filtration. T means threshold. The results are on VisDrone2019.}

   \scalebox{1}{
\begin{tabular}{c|cc|ccc}
    \toprule
  \setrowcolor  Method & AP & AP$_{0.5}$ & AP$_{vt}$ & AP$_t$ & AP$_s$ \\
    \midrule
    w/o T & 27.0 & 46.6 & 2.2 & 11.3 & 24.2 \\
    Fixed T & 27.1 & 46.8 & \textbf{2.3} & 11.2 & 24.0 \\
    Learnable T & \textbf{27.3} & \textbf{46.9} & \textbf{2.3} & \textbf{11.5} & \textbf{24.7} \\
  \bottomrule
  \end{tabular}}
 \label{tab:ablation_3}
\end{minipage}
\begin{minipage}[c]{0.49\textwidth}
\centering
   \captionof{table}{Effectiveness of different feature enhancement methods. EA means element-wise attention, Concat means concatenate and EM means element-wise multiplication.}
   \scalebox{1.05}{
\begin{tabular}{c|cc|ccc}
    \toprule
  \setrowcolor  Method & AP & AP$_{0.5}$ & AP$_{vt}$ & AP$_t$ & AP$_s$ \\
    \midrule
     EM & 36.2 & 84.9 & 28.8 & 45.7 & 58.4 \\
     Concat & 36.6 & 85.3 & 29.0 & 46.0 & 58.7 \\
     EA & \textbf{38.3} & \textbf{87.4} & \textbf{30.8} & \textbf{47.4} & \textbf{59.4} \\
  \bottomrule
  \end{tabular}}
 \label{tab:ablation_2}
\end{minipage}
\begin{minipage}[c]{0.49\textwidth}
\centering

   \captionof{table}{Performance of different designs of difference map. PD means the pixel difference map and HFD means the high-frequency difference map.}

   \scalebox{1.05}{
\begin{tabular}{c|cc|ccc}
    \toprule
  \setrowcolor  Method & AP & AP$_{0.5}$ & AP$_{vt}$ & AP$_t$ & AP$_s$ \\
    \midrule
    baseline & 36.4 & 85.0 & 28.8 & 45.7 & 58.3 \\
    PD & 38.3 & 87.4 & 30.8 & 47.4 & \textbf{59.4} \\
    HFD & \textbf{38.4} & \textbf{87.6} & \textbf{31.1} & \textbf{47.6} & 59.2 \\
  \bottomrule
  \end{tabular}}
 \label{tab:ablation_4}
\end{minipage}
\end{table}

\subsection{Ablation Study and Discussion}

\textbf{Effectiveness of individual component.}
As shown in Sec.\ref{sec3.2} and Sec.\ref{sec3.3}, the self-reconstruction mechanism consists of two key components: the Reconstruction Head and the Difference Map Guided Feature Enhancement module.
Note that the validation of DGFE relies on the Reconstruction Head, as it requires the construction of a difference map as input.
When the reconstruction head is applied alone, we obtain the reconstructed image without constructing a difference map.
We add Reconstruction Head and DGFE into the Cascade R-CNN one-by-one.
The results are listed in Tab. \ref{tab:ablation_1}.
The DGFE increased 1.8 points in AP over the use of the reconstruction head alone, accompanied by a substantial gain of 2.1 points in AP$_{vt}$.
These results conclusively prove that DGFE is both an effective and indispensable module for leveraging the difference map.
It should be highlighted that the sole application of the reconstruction head yielded a modest improvement of 0.1 AP, along with an increment of 0.2 points in both AP$_t$ and AP$_s$.
This enhancement can be attributed to the fact that integrating image reconstruction constraints within the object detection model not only preserves detection performance but also, to a certain degree, improves the backbone network's pixel comprehension.
This approach could potentially offer fresh inspiration for the field of object detection.
\begin{wrapfigure}[12]{r}{0.5\textwidth}
    \includegraphics[width=1\linewidth]{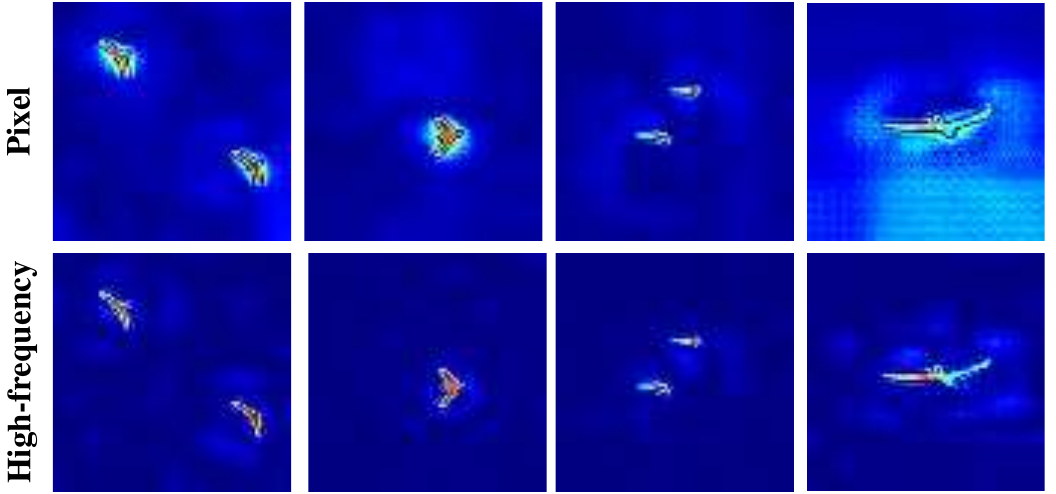}
    \caption{Visualizations of the pixel difference maps and high-frequency difference maps.}
    \label{fig:vis_high_frequency}
\end{wrapfigure}
\begin{wrapfigure}[16]{r}{0.5\textwidth}
    \includegraphics[width=1\linewidth]{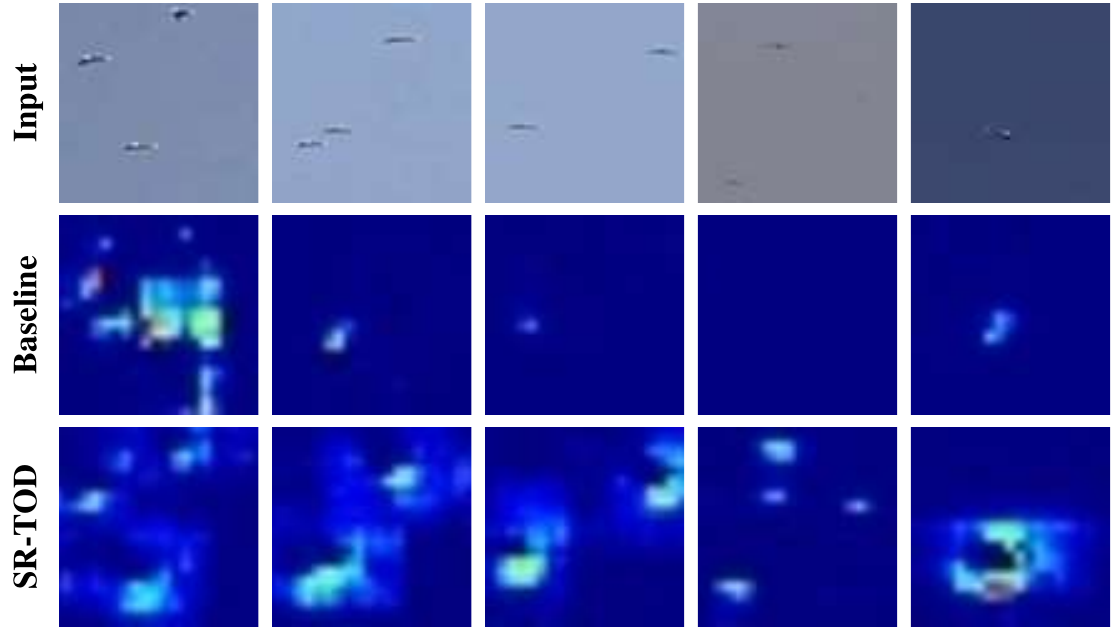}
 \caption{Visualizations on DroneSwarms. The first row is the local regions of input images. The second and third rows are the corresponding feature maps from Cascade R-CNN $w/o$ and $w/$ SR-TOD.}
    \label{fig:vis_analyse}
\end{wrapfigure}

\noindent\textbf{Effectiveness of different feature enhancement methods.}
In this section, we explore various feature enhancement techniques using the Cascade R-CNN framework to incorporate prior knowledge from the difference map.
Beyond employing element-wise attention, we experiment with fusion method that concatenate the binary difference map with the feature map and explore approach that execute direct element-wise multiplication between the two.
As shown in Tab. \ref{tab:ablation_2}, it is evident that element-wise attention markedly surpasses the performance of alternative methods.
The fusion method using the concatenate operation has a weak effect, as direct fusion brings more noise and diminishes the spatial positional information of the difference map.
The approach that employs element-wise multiplication is deemed too simplistic and basic, as numerous inactive areas in the difference map can cause the obliteration of original features in the feature map, ultimately impairing detection capabilities. Consequently, we opt for element-wise attention in DGFE, which provides a more sophisticated solution.

\noindent\textbf{Effectiveness of threshold filtration.}
To investigate the effect of thresholds, we also explore methods employing fixed thresholds and those without threshold-based filtering.
We conduct experiments on the VisDrone2019 dataset~\cite{du2019visdrone}, which features intricate backgrounds and diverse categories.
The results are shown in Tab. \ref{tab:ablation_3}.
Undoubtedly, the approaches that involve setting a threshold generally exceed the performance of the method without a threshold, thus affirming the utility of thresholding.
In addition, the learnable threshold method performs similarly to the fixed threshold method in AP$_{vt}$ but shows a 0.3 points improvement in AP$_t$ and a 0.7 points improvement in AP$_s$.
This indicates that tiny objects with a size above 8 pixels are more sensitive to thresholds.
Therefore, the method with a learnable threshold achieves better performance by finely adjusting the threshold.
\\ \textbf{Exploration of the high-frequency difference map.}
Recently, some image reconstruction methods~\cite{zhou2020guided,jiang2021focal} model images in the frequency domain.
The high-frequency information of an image includes details such as edges and textures, which are precisely the parts that are difficult to recover in image self-reconstruction. In particular, noise is also considered part of high-frequency information, and very tiny objects can be approximated as noise-like points with a size of only a few pixels.
Inspired by this, we further utilize the fast Fourier transform algorithm to separate the high-frequency components of the original input image and the reconstructed image, and construct high-frequency difference maps, as shown in Fig. \ref{fig:vis_high_frequency}.
Although the high-frequency difference map and pixel difference map are quite similar, the high-frequency difference map has more refined contours on some drone objects.
Additionally, the high-frequency difference map significantly reduces noise signals.
However, this also results in certain smaller drone objects appearing even more faint in the high-frequency difference map.
Then we replace the pixel difference map (PD) with the high-frequency difference map (HFD) and conduct experiments on DroneSwarms using Cascade R-CNN w/ SR-TOD.
The results are shown in Tab. \ref{tab:ablation_4}.
The HFD has a 0.1 AP improvement compared to PD.
Moreover, there is an improvement of 0.3 points in AP$_{vt}$ and 0.2 points in AP$_t$. This demonstrates that the high-frequency difference map can provide more accurate priors for tiny objects compared to the pixel difference map.
This indicates that the information severely lost in tiny object feature extraction mainly lies in the high-frequency components.
Note that compared to the baseline, the detection performance of using pixel difference map and high-frequency difference map is very similar.
Considering computational efficiency, we use the pixel difference map as the basic setting for SR-TOD.
However, the high-frequency difference map shows the potential for further exploration of our method.
More results are provided in the supplementary materials.

\subsection{Visualization Analysis}

To demonstrate the effect of DGFE more clearly and intuitively, we perform visualizations on DroneSwarms.
The results are shown in Fig.~\ref{fig:vis_analyse}.
From the results in the second row, it can be seen that before feature enhancement, the detection model lacks sufficient attention to tiny drones.
The signals of some tiny objects are very weak and almost wiped out, resulting in invisibility.
The results in the third row show that the SR-TOD significantly enhances the feature representation of tiny objects, which makes the tiny objects visible and clear to the detectors.

\section{Conclusion}

In this paper, we analyze the challenge of information loss in tiny object detection and the limitations faced by generative methods attempting to alleviate this issue.
To this end, we introduce an image self-reconstruction mechanism, constructing difference maps as the prior information of tiny objects, making the feature more visible to detectors.
Then, we further design a Difference Map Guided Feature Enhancement (DGFE) module to improve the feature representation of tiny objects, providing more clear representations.
Experiments on our proposed DroneSwarms and two other datasets show the superiority and robustness of the SR-TOD. In the future, we will explore more effective ways to construct more accurate difference maps for tiny objects.

\section*{Acknowledgements}
This work was sponsored in part by the National Key R\&D Program of China 2022ZD0116500, in part by the National Natural Science Foundation of China (62106171, 62222608, U23B2049, 61925602), in part by the Haihe Lab of ITAI under Grant 22HHXCJC00002, in part by Tianjin Natural Science Funds for Distinguished Young Scholar under Grant 23JCJQJC00270, and in part by the Key Laboratory of Big Data Intelligent Computing, Chongqing University of Posts and Telecommunications under Grant BDIC-2023-A-008. This work was also sponsored by CAAI-CANN Open Fund, developed on OpenI Community. 


%
%
\bibliographystyle{splncs04}
\bibliography{CameraReady}
\end{document}